\documentclass{article}
\usepackage{spconf,amsmath,graphicx}
\usepackage{bm}
\usepackage{color}
\usepackage{url}
\usepackage{fancyvrb}
\usepackage[subrefformat=parens]{subcaption}

\title{The pipeline system of ASR and NLU with MLM-based data augmentation \\ toward STOP low-resource challenge}

\name{
\begin{tabular}{c}
Hayato Futami$^1$, Jessica Huynh$^2$, Siddhant Arora$^2$, Shih-Lun Wu$^2$, Yosuke Kashiwagi$^1$, \\ Yifan Peng$^2$, Brian Yan$^2$, Emiru Tsunoo$^1$, Shinji Watanabe$^2$
\end{tabular}
}
\address{$^1$Sony Group Corporation, Japan $^2$Carnegie Mellon University, USA}

\begin{document}
%
\maketitle
%

\begin{abstract}
This paper describes our system for the low-resource domain adaptation track (Track 3) in Spoken Language Understanding Grand Challenge, which is a part of ICASSP Signal Processing Grand Challenge 2023.
In the track, we adopt a pipeline approach of ASR and NLU.
For ASR, we fine-tune Whisper for each domain with upsampling.
For NLU, we fine-tune BART on all the Track3 data and then on low-resource domain data.
We apply masked LM (MLM) -based data augmentation, where some of input tokens and corresponding target labels are replaced using MLM.
We also apply a retrieval-based approach, where model input is augmented with similar training samples.
As a result, we achieved exact match (EM) accuracy $63.3/75.0$ (average: $69.15$) for reminder/weather domain, and won {\bf the 1st place} at the challenge.
\end{abstract}

\vspace{-10pt}
\section{Introduction}
\vspace{-5pt}
In the Spoken Language Understanding Grand Challenge at ICASSP 2023, we aim to predict user's intents, slots types and values from audio, known as spoken language understanding (SLU), using the STOP \cite{Tomasello22-STOP} dataset.
The dataset is created by adding audio recordings to TOPv2 \cite{Chen20-TOPv2} dataset.
It consists of 8 domains: alarm, event, messaging, music, navigation, timer, reminder, and weather.
For Track 3, we address low-resource domain adaptation.
The number of training samples for reminder/weather are limited to 482/162, which ensures at least 25 samples are included for each intent and slot type.
We can use held-in (6 domains) and low-resource data (reminder/weather), which we call "Track3 data".

We adopt a pipeline approach for Track 3.
An ASR model generates transcripts from audio, and then an NLU model converts transcripts into semantic parse results.
This apporach fully benefits from both powerful pre-trained ASR and LM, which is especially important in this low-resource setting.
For ASR, we fine-tuned Whisper \cite{Radford22-Whisper} on the STOP dataset for Track 3.
We built two individual models for each target domain: one was fine-tuned on held-in and reminder data, and the other was fine-tuned on held-in and weather data.
As low-resource data was much smaller than held-in data, they were upsampled by $20$ times.
For NLU, we applied two-step fine-tuning.
First, we fine-tuned BART on all the Track3 data.
Then, we fine-tuned it on each low-resource domain (reminder/weather) data to build two models.
During low-resource fine-tuning, we applied masked LM (MLM) -based data augmentation.
We also retrieved related training samples for an additional model input, which is known as retrieval augmentation \cite{Zemlyanskiy22-GR}.

\vspace{-10pt}
\section{Augmentations for NLU}
\vspace{-5pt}
\subsection{MLM-based data augmentation}
\label{section:da}
Since low-resource sets have limited samples for training, its diversity would not be enough and prone to overfit.
To solve the issue, we apply masked LM (MLM) -based data augmentation.
Let $\bm{x}$ denote an input transcript and $\bm{y}$ denote a target semantic parse.
The data augmentation procedure is:
\begin{enumerate}
    \item Some portion ($p \sim U(0, 0.2)$) of tokens in $\bm{x}$ are randomly masked to make $\bm{x}_{\rm mask}$.
    \item The masked tokens are replaced with MLM's predictions to make $\bm{x}_{\rm aug}$.
    \item In case the original tokens before masking exist in $\bm{y}$ (as slot values), they are also replaced to make $\bm{y}_{\rm aug}$.
\end{enumerate}
Table \ref{tab:example-generation} shows an example of data augmentation.
We sometimes get ($\bm{x}_{\rm aug}$, $\bm{y}_{\rm aug}$) that is grammatically incorrect or does not fit to intent and slot types.
We filter out samples that cannot be exactly inferred by an NLU model once trained on the original data.
Similar data augmentation is done at \cite{Tran20-GS}, but ours does not require any fine-tuning for a generator MLM.

\begingroup
\renewcommand{\arraystretch}{1.1}
\begin{table}[t]
  \caption{Example of data augmentation.}
  \label{tab:example-generation}
  \centering
  \begin{tabular}{ll} \hline
    $\bm{x}$: & how ' s the weather in sydney \\
    $\bm{y}$: & [in:get\_weather [sl:location sydney ] ] \\
    $\bm{x}_{\rm mask}$: & how ' s the weather in \url{[MASK]} \\
    $\bm{x}_{\rm aug}$: & how ' s the weather in {\bf london} \\
    $\bm{y}_{\rm aug}$: & [in:get\_weather [sl:location {\bf london} ] ] \\ \hline
 \end{tabular}
\end{table}
\endgroup

\vspace{-5pt}
\subsection{Retrieval augmentation}
\label{section:ra}
We follow \cite{Zemlyanskiy22-GR}, which shows the effectiveness of retrieval augmentation in low-resource domain adaptation on TOPv2 dataset.
We add $k=4$ input--output pairs as examplars to $\bm{x}$:
\begin{align*}
\bm{x}' = \bm{x} \, ; \, \bm{x}_1 \, ; \, \bm{y}_1 \, ; \, ... \, ; \, \bm{x}_4 \, ; \, \bm{y}_4 
\end{align*}
These examplars are selected from Track3 training data, based on TF-IDF similarity score.
We only consider input similarity between $\bm{x}$ and $\bm{x}_i$, for simplicity.
During training, examplars are sampled over geometric distribution, where $r$-ranked samples are selected with a probability of $p(1-p)^{r-1}$ ($p=0.1$).

\vspace{-10pt}
\section{Experimental Evaluations}
\vspace{-5pt}
For ASR, we fine-tuned Whisper model \footnote{https://huggingface.co/openai/whisper-medium} using ESPnet \footnote{https://github.com/espnet/espnet} framework.
We fine-tuned it for each domain, on the mixture of held-in data and $20$x upsampled low-resource domain data.
We applied speed perturbation, SpecAugment, and label smoothing ($p=0.1$).
We averaged $10$-best checkpoints based on validation set.
Table \ref{tab:asr} shows the ASR results.
With fine-tuning (FT), the WERs were observed to be improved.
All the words were lowercased in both ASR and NLU.

\begingroup
\renewcommand{\arraystretch}{1.1}
\begin{table}[t]
  \caption{ASR performance of STOP dataset.}
  \label{tab:asr}
  \centering
  \begin{tabular}{lcc} \hline
     & \multicolumn{2}{c}{Valid/Test WER[\%]} \\
     & reminder & weather \\ \hline
    Whisper & $2.5/3.7$ & $4.1/2.9$ \\
    + FT & $1.7/2.8$ & $3.4/2.3$ \\ \hline
 \end{tabular}
\end{table}
\endgroup

For NLU, we fine-tuned BART model \footnote{https://huggingface.co/facebook/bart-large} using Transformers \footnote{https://github.com/huggingface/transformers} framework.
To solve NLU, intent and slot tags (e.g. \url{[in:get_weather}) were added to the original BART vocabulary.
Table \ref{tab:nlu} shows the NLU results, where all the evaluation is done using ground-truth text as model input.
We first fine-tuned BART on all the Track3 data, and then fine-tuned it on low-resource data, which we call low-resource fine-tuning (LR-FT).
By adding LR-FT, the EM (exact match) accuracy was improved.
As described in Section \ref{section:da}, we applied data augmentation with BERT \footnote{https://huggingface.co/bert-base-uncased} during LR-FT.
We prepared $3241/881$ synthetic samples, and $1841/530$ samples remained after filtering.
We found data augmentation improved the EM accuracy.
We further applied retrieval augmentation to BART (RA-BART), as described in Section \ref{section:ra}.
They are also fine-tuned in two steps (all + reminder/weather).
With combination of data augmentation and retrieval, the EM score was further improved to $73.8/79.7$ (average: $76.8$).
Note that we must apply the same methodology for both low-resource domains, so we selected the model of the best averaged EM accuracy on validation set.

\begingroup
\renewcommand{\arraystretch}{1.1}
\begin{table}[t]
  \caption{NLU performance from {\bf ground-truth} text.}
  \label{tab:nlu}
  \centering
  \begin{tabular}{lcc} \hline
     & \multicolumn{2}{c}{Valid/Test EM Acc.[\%]} \\
     & reminder & weather \\ \hline
    BART & $41.3/45.1$ & $39.1/65.1$ \\
    + LR-FT & $68.2/67.5$ & $76.7/79.7$ \\
    + LR-FT (dataaug) & $68.2/69.9$ & $78.9/80.3$ \\ \hline
    RA-BART +LR-FT & $66.7/70.3$ & $81.2/80.3$ \\
    + LR-FT (dataaug) & $69.7/73.8$ & $80.5/79.7$ \\ \hline
 \end{tabular}
\end{table}
\endgroup

Our target is to predict semantic parse from audio.
Table \ref{tab:slu} shows the end-to-end SLU evaluation with our ASR and NLU.
We achieved the EM accuracy $\bm{63.3}/\bm{75.0}$ (average: $\bm{69.15}$).
Our pipeline was confirmed to be much better than the E2E SLU baseline \cite{Tomasello22-STOP}.

\begingroup
\renewcommand{\arraystretch}{1.1}
\begin{table}[t]
  \caption{SLU evaluation from {\bf audio}, which is the target of the challenge.}
  \label{tab:slu}
  \centering
  \begin{tabular}{lccc} \hline
     & \multicolumn{3}{c}{Test EM Acc.[\%]} \\
     & reminder & weather & Avg. \\ \hline
    Our pipeline & $63.3$ & $75.0$ & $69.15$ \\ \hline
    E2E SLU \cite{Tomasello22-STOP} & $15.38$ & $46.77$ & $31.08$ \\ \hline
 \end{tabular}
\end{table}
\endgroup

\section{Acknowledgement}   
\vspace{-5pt}
This work used Bridges2 system at PSC and Delta system at NCSA through allocation CIS210014 from the Advanced Cyberinfrastructure Coordination Ecosystem: Services \& Support (ACCESS) program, which is supported by National Science Foundation grants \#2138259, \#2138286, \#2138307, \#2137603, and \#2138296.
Jessica Huynh was supported by the NSF Graduate Research Fellowship under Grant Nos. DGE1745016 and DGE2140739. The opinions expressed in this paper do not necessarily reflect those of that funding agency.

\vspace{-10pt}

\bibliographystyle{IEEEbib}
\bibliography{strings,refs}

\begin{thebibliography}{1}

\bibitem{Tomasello22-STOP}
Paden Tomasello et~al.,
\newblock ``{STOP}: A dataset for spoken task oriented semantic parsing,''
\newblock {\em SLT}, pp. 991--998, 2022.

\bibitem{Chen20-TOPv2}
Xilun Chen et~al.,
\newblock ``Low-resource domain adaptation for compositional task-oriented
  semantic parsing,''
\newblock in {\em EMNLP}, 2020, pp. 5090--5100.

\bibitem{Radford22-Whisper}
Alec Radford et~al.,
\newblock ``Robust speech recognition via large-scale weak supervision,''
\newblock {\em arXiv}, 2022.

\bibitem{Zemlyanskiy22-GR}
Yury Zemlyanskiy et~al.,
\newblock ``Generate-and-retrieve: Use your predictions to improve retrieval
  for semantic parsing,''
\newblock in {\em COLING}, 2022, pp. 4946--4951.

\bibitem{Tran20-GS}
Ke~Tran et~al.,
\newblock ``Generating synthetic data for task-oriented semantic parsing with
  hierarchical representations,''
\newblock in {\em SPNLP}, 2020, pp. 17--21.

\end{thebibliography}

\end{document}